\documentclass[10pt, a4paper, conference]{IEEEtran}

\IEEEoverridecommandlockouts
\usepackage{cite}
\usepackage{amsmath,amssymb,amsfonts}
\usepackage{algorithmic}
\usepackage{graphicx}
\usepackage{textcomp}
\usepackage{xcolor}
\def\BibTeX{{\rm B\kern-.05em{\sc i\kern-.025em b}\kern-.08em
    T\kern-.1667em\lower.7ex\hbox{E}\kern-.125emX}}
\begin{document}

\title{Seed Kernel Counting using Deep Learning for High-Throughput Phenotyping \\}

\author{\IEEEauthorblockN{Venkat Margapuri} \\
\IEEEauthorblockA{\textit{Department of Computer Science} \\
\textit{Villanova University}\\
Villanova, PA \\
vmargapu@villanova.edu}\\

\and
\IEEEauthorblockN{Prapti Thapaliya} \\
\IEEEauthorblockA{\textit{Department of Computer Science} \\
\textit{Villanova University}\\
Villanova, PA \\
pthapali@villanova.edu}\\
\and
\IEEEauthorblockN{Mitchell Neilsen} \\
\IEEEauthorblockA{\textit{Department of Computer Science} \\
\textit{Kansas State University}\\
Manhattan, KS \\
neilsen@ksu.edu}
}

\maketitle

\begin{abstract}
High-throughput phenotyping (HTP) of seeds, also known as seed phenotyping, is the comprehensive assessment of complex seed traits such as growth, development, tolerance, resistance, ecology, yield, and the measurement of parameters that form more complex traits \cite{b0}. One of the key aspects of seed phenotyping is cereal yield estimation that the seed packaging industry relies upon to conduct their business. While mechanized seed kernel counters are available in the market, they are often priced high and outside the affordability realm of small-scale seed packaging firms. The development of object tracking neural network models such as You Only Look Once (YOLO) enables computer scientists to design algorithms that can estimate cereal yield inexpensively. The key bottleneck of neural network models is that they require a plethora of labeled training data before they can be put to task. We demonstrate that the use of synthetic imagery serves as a feasible substitute to train neural networks for object tracking that includes the tasks of object classification and detection. Furthermore, we propose a seed kernel counter that uses a low-cost mechanical hopper, trained YOLOv8 neural network model, and object tracking algorithms of StrongSORT and ByteTrack to estimate cereal yield from videos captured using a mobile phone. The experiment yields a seed kernel count with an accuracy of 95.2\% and 93.2\% for Soy and Wheat respectively using the StrongSORT algorithm, and an accuracy of 96.8\% and 92.4\% for Soy and Wheat respectively using the ByteTrack algorithm.
 
\end{abstract}


\section{Introduction}
The advent of technology in the field of agriculture commenced over a century ago, and several studies have been conducted since the 1990s to improve production efficiency \cite{b1}. High-throughput Phenotyping (HTP) of seeds, also known as seed phenotyping, is the comprehensive assessment of complex seed traits such as growth, development, tolerance, resistance, ecology, yield, and the measurement of parameters that form more complex traits \cite{b0}. HTP increases the accuracy of measurements while reducing costs through the application of automation, remote sensing, data integration, and experimental design. The current work addresses the aspect of cereal yield estimation, a use case primarily geared toward the seed packaging industry. In the absence of accurate seed kernel count, seed packaging firms package bags of seed kernels by weight. However, the weight of seed kernels is influenced by the environment in which the seed kernel resides. For instance, the weight of the seed kernels increases when they are soaked in water from the moisture they absorb. There is often a discrepancy between the expected and actual number of seed kernels when the customer acquires the seed kernel bags packaged by weight. Packaging seeds by count is tedious when performed by hand, making it infeasible. Seed packaging firms have to resort to the use of expensive mechanized seed counting machinery to pack seed kernels by count. However, the seed counting machinery, ranging from \$500 to \$3000, is overly expensive for small-scale packaging firms that operate on a tight budget. Furthermore, it deters them from the ability to pack by count, negatively impacting their customers. \newline
\null \quad The significant developments made in the field of artificial neural networks, specifically object tracking neural network models, enable plant and computer scientists to collaboratively develop low-cost, high-throughput systems for seed kernel counting. This paper demonstrates the idea of leveraging seed kernel videos for soy and wheat to estimate seed kernel count using low-cost hardware components and the object tracking neural network model of You Only Look Once (YOLO). YOLO belongs to a class of supervised neural networks that provides object tracking ability. The task of object tracking bundles the tasks of object detection and classification within it. Supervised neural network models require a plethora of labeled information to train for tasks such as object detection and classification. However, large amounts of labeled training data are not always readily available for entities such as seed kernels. This paper demonstrates that the use of synthetic image datasets generated following the principles of Domain Randomization (DR) \cite{b2, b3, b4} is a feasible alternative to train neural network models in the absence of real-world labeled datasets. \newline
\null \quad The key contributions of the paper are:\begin{enumerate}
    \item The assembly of relatively inexpensive hardware components to capture the videos of wheat and soy seed kernels for deep learning in the domain of agriculture.
    \item The application of DR to generate synthetic image datasets for deep learning applications on wheat and soy seed kernels.   
    \item The use of object tracking neural network models to detect and count seed kernels from videos as an alternative to the relatively expensive commercial seed counters available in the market.
    
\end{enumerate}

\section{Related Work}

Neilsen et al. \cite{b5} proposed an image processing algorithm to perform seed kernel counting from videos. The working of the algorithm is based on tracking each of the seed kernels as they flow down a backlit platform. A seed kernel is considered a valid detection and counted if the seed kernel is detected a predefined number of times (threshold). However, the image processing algorithm is highly sensitive to the video's frame rate. \newline
\null \quad GridFree \cite{b6} is a Python package for image analysis of interactive grain counting and measurement. GridFree uses an unsupervised machine learning approach, K-Means, to segment kernels from the background by using principal component analysis (PCA) on both raw image channels and their color indices. The package incorporates users’ experiences as a dynamic criterion to set thresholds for a divide-and-combine strategy that effectively segments adjacent kernels. When adjacent multiple kernels are incorrectly segmented as a single object, they form an outlier on the distribution plot of kernel area, length, and width. The software exhibits great performance on multiple crop types such as alfalfa, canola, lentil, wheat, chickpea, and soybean. \newline
\null \quad Parico et al. \cite{b7} performed real-time pear fruit detection and counting using YOLOv4 models and Deep SORT algorithm. The study provides a systematic and pragmatic methodology to choose the most suitable neural network model for a desired application in the field of agriculture. The region-of-interest (ROI) line technique was used by the study to estimate the number of pear fruits detected by the neural network model. \newline
\null \quad Wu et. al. \cite{b8} performed detection of Camellia oleifera fruit in complex scenes by using YOLOv7 and data augmentation. The comparison of YOLOv7's performance with YOLOv5, YOLOv3-spp, and Faster RCNN showed that YOLOv7 provided the best detection performance. The experiment yielded a Mean Average Precision, Precision, Recall, F1 Score, and average detection time of 96.03\%, 94.76\%, 95.54\%, 95.15\%, and 0.025 seconds per image respectively. \newline
\null \quad Hajar et al. \cite{b9} performed vision-based moving obstacle detection and tracking in paddy field using YOLOv3 and Deep SORT. The center point positions of the obstacles were used to track the objects as they moved through the paddy field. The augmented YOLOv3 architecture consisted of 23 residual blocks and up-sampled only once. The augmented architecture obtained a mean intersection over union score of 0.779 and was 27.3\% faster in processing speed than standard YOLOv3. \newline
\null \quad Huang et al. \cite{b10} developed a video-based detection model to identify damage within unwashed eggs using YOLOv5 and ByteTrack object tracking algorithm. The detection results of the different frames were associated by ID, and used five different frames to determine egg category. The experimental results showed an accuracy of 96.4\% when YOLOv5 in conjunction with the ByteTrack algorithm was used to detect broken/damaged eggs from videos.  

\section{Hardware Components}
\label{HardwareComponents}
\begin{figure}[h!]
\centerline{\includegraphics{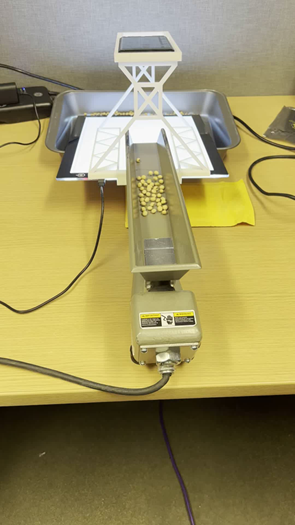}}
\caption{Mechanical hopper delivering seed kernels}
\label{MechanicalHopper}
\end{figure}

This article proposes a low-cost setup for the capture of seed kernel videos to aid in seed kernel counting. Fig. ~\ref{MechanicalHopper} shows the seed kernel image capture setup designed for the experiment wherein a mechanical hopper delivers seed kernels onto a backlit platform. 

\begin{figure}[h]
\centerline{\includegraphics[scale=0.75]{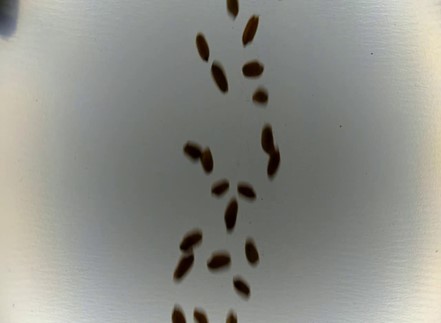}}
\caption{Wheat seed kernels flowing down the light box}
\label{WheatSeeds}
\end{figure}

The mechanical hopper helps to deliver seeds at a constant rate, unlike free-hand delivery which tends to be erratic. A mobile phone is placed on a 3-D printed stand and held on the backlit platform. The 3-D printed stand which holds the mobile phone ensures that the camera (mobile phone) is always held orthogonal to the surface, thereby, eliminating any skew that may result during the capture of the video. Furthermore, the stand is fitted with 3-D printed rails at the bottom to channel the seed kernels and ensure that they remain in the field of view of the camera as they roll down the backlit platform. In the absence of the rails, it is observed that seed kernels often drift to the side and fall off the backlit platform prematurely, hindering their detection. The mobile phone used for image capture is a Google Pixel 2 XL mobile phone whose default capture frame rate is 60 fps. Fig. ~\ref{WheatSeeds} shows a frame of the wheat seed kernel video captured using the proposed setup in Fig. ~\ref{MechanicalHopper}. \newline

\section{Domain Randomization and Image Datasets}
\label{DR}
\begin{figure}[htbp]
\centerline{\includegraphics[scale=0.75]{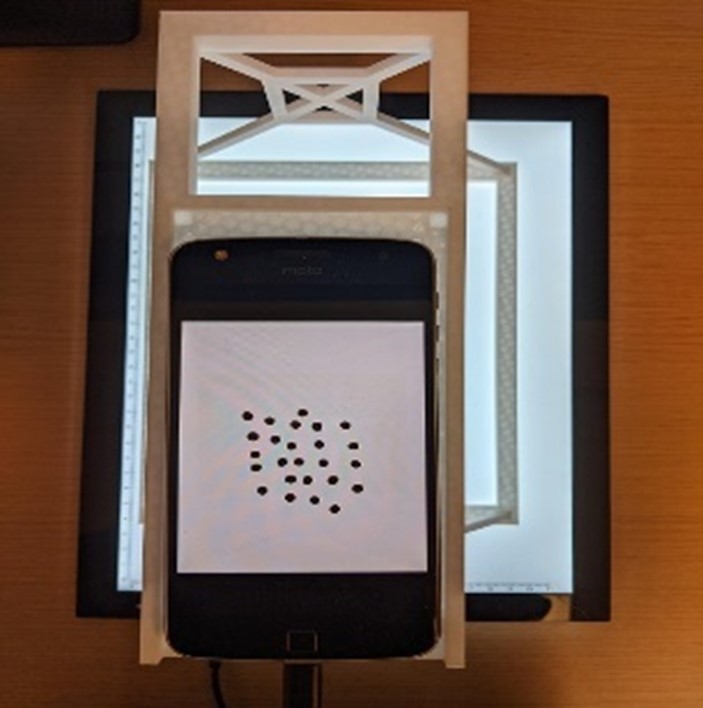}}
\caption{Image capture of soy seed kernels}
\label{soy kernel}
\end{figure}

Domain Randomization (DR) is the idea of training neural network models on images containing simulated objects that translate closely to real-world objects. A small sample of images containing real objects is required for the creation of synthetic images using DR. The images of soy and wheat are captured by placing a mobile phone on a 3-D printed stand that holds the mobile phone orthogonal to the surface. Fig. ~\ref{soy kernel} shows soy seeds being captured by the proposed image capture setup.

\begin{figure}[htbp]
\centerline{\includegraphics[scale=0.75]{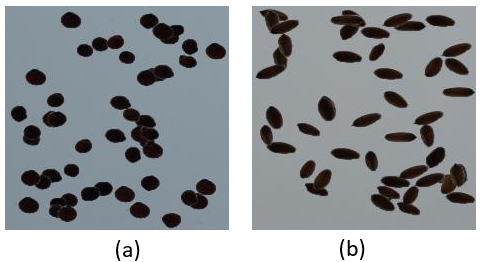}}
\caption{Sample images of (a) soy (b) wheat from the synthetic image dataset }
\label{soyandwheat}
\end{figure}

\null \quad Images of 25 seed kernels of soy and wheat are captured using the setup shown in Fig. ~\ref{soy kernel}. Using the synthetic image generator developed as part of the authors' previous work \cite{b2}, synthetic images containing seed kernels of soy and wheat are generated. The steps in the application of DR to create the synthetic image datasets are briefly outlined below. \begin{enumerate}
    \item Capture images of each of the seed kernels and extract the seed kernels from the image by extracting the alpha channel of the image which corresponds to the foreground.
    \item Lay each foreground (seed kernels) on a background canvas which ensures that the foreground is clearly visible. It is recommended to apply different augmentations to the foreground as it is overlaid on the background.
\end{enumerate}
The synthetic image generator employed for this work applies image augmentation techniques such as rotation, flipping, and noising to generate image datasets that are akin to real-world scenarios. The synthetic images are meticulously curated to allow for about 25\% overlap at the maximum to account for clustered seed kernels as the frames of the video are processed. Synthetic image datasets are created for the seed types of soy and wheat, wherein each dataset consists of 200 images of size 320x320x3 with each image containing between 25 and 50 seed kernels overlaid on a light background, as shown in Fig. ~\ref{soyandwheat}. Furthermore, the synthetic image generator outputs annotation files that contain location coordinates pertinent to each seed kernel in the image in the TXT format for YOLOv8 to consume and process during training.

\section{YOLOv8 and Object Tracking Algorithms}
\label{YOLO}
The YOLO model is a single-shot detector \cite{b11, b12, b13} that uses a fully convolutional neural network as the backbone to process the input image. The YOLOv8 \cite{b14} model used for this work was released in January 2023 by Ultralytics to support multiple computer vision tasks such as object detection, classification, segmentation, pose estimation, and tracking. YOLOv8 comprises a convolutional neural network that is divided into two parts: backbone and head. The backbone, CSPDarknet53 \cite{b15}, consists of 53 convolutional layers and uses cross state partial connections to facilitate information flow between different layers. The head consists of multiple convolutional layers followed by a series of fully connected layers. Another key characteristic of YOLOv8 is the use of the Self-Attention \cite{b16} \cite{b17} mechanism in the head to enable the model to focus on specific parts of the image for better performance. \newline
\null \quad Object tracking involves the tasks of object detection and classification, and feature mapping between objects in different frames as they move around. YOLOv8 is a single-shot object detector that detects and labels objects in a given image. However, object tracking requires that the object be detected in every frame across the video i.e. the object is to be re-identified within every frame as each frame of the video is processed. Several object tracking algorithms \cite{b18, b19, b20} have been proposed over the years. This paper considers two object tracking algorithms for experimentation, StrongSORT and ByteTrack. The StrongSORT and ByteTrack algorithms are selected for their ability to handle complete and partial occlusions. Furthermore, both algorithms demonstrate acute ability to track moving objects in real-time. 
\null \subsection{StrongSORT}
The StrongSORT \cite{b18} algorithm is an improvement over the DeepSORT \cite{b19} algorithm. StrongSORT algorithm, like DeepSORT, is a two-branch framework consisting of an Appearance branch and a Motion branch. The algorithm works on objects that have been detected in each frame of the video. The YOLOX-X \cite{b20} model is used to detect the objects in the frame. However, detections may be performed using alternate object detectors such as Faster R-CNN \cite{b23} as well. \newline
\null \quad The Appearance branch identifies the features of each of the objects detected in a given frame. The detected features are used to match a given object across different frames. BoT \cite{b21} is leveraged as the feature extractor by the StrongSORT algorithm. The appearance state for the $i^{th}$ tracklet within frame t, $e_{i}^{t}$, is updated as the exponential moving average (EMA) given by $e_{i}^{t}$ = $\alpha$$e_{i}^{t-1}$ + (1-$\alpha$)$f_{i}^{t}$ where $f_{i}^{t}$ is the appearance embedding of the current matched detection and $\alpha$ = 0.9, is a variable momentum term. The benefit of EMA lies in the fact that it leverages information of inter-frame feature changes and suppresses detection noise.  \newline
\null \quad The Motion branch leverages Kalman Filter \cite{b24} to predict the position of the object in the frame based on a constant velocity model. However, the basic Kalman Filter offers poor performance when the detections made by the object detector are sub par. In order to tackle this issue, the StrongSORT algorithm uses the NSA Kalman Filter algorithm borrowed from the GIAO tracker \cite{b25}. 
\subsection{ByteTrack}
ByteTrack \cite{b22} algorithm is best suited to cases where objects detected in one frame of the video by detectors such as YOLOv8 are ignored in a different frame of the video. Such instances are generally a result of occlusion of objects as they move from one frame to the other. Motion tracking (MoT) algorithms typically leverage bounding boxes plotted by object detectors to assign a unique ID to each object in the video. Furthermore, object detectors associate a confidence level to each bounding box that is plotted. Most MoT algorithms ignore bounding boxes that have a confidence level below a threshold to avoid false positives. However, ignoring the bounding boxes at low confidence levels risks ignoring detecting objects that may otherwise be detected (true positives). The ByteTrack algorithm mitigates the risk by leveraging bounding boxes at all confidence levels agnostic of a confidence level threshold and meticulously attempts to identify all objects in a frame agnostic of the detection confidence level. \newline
\null \quad The ByteTrack algorithm leverages a queue called Tracklets to store all the objects (and bounding boxes) that have been detected by the object detector. The bounding boxes are separated into high score ($D^{high}$) and low score ($D^{low}$) based on threshold. Each of the objects in the Tracklets queue is tracked across each frame of the video agnostic of the confidence level of the bounding boxes associated to them. The tracking of objects across frames is performed using Kalman Filter \cite{b24}. Firstly, the position (bounding box) of each of the objects in the tracklets queue is predicted in the subsequent frame. The predictions are matched with the actual detections made by the object detector using Motion Similarity score. Motion Similarity score is computed as the result of Intersection over Union (IoU) between the predicted and actual bounding boxes. Initially, tracklet matching is done between the predicted and high score ($D^{high}$) bounding boxes. The tracklets that do not match with any of the high score bounding boxes are matched with low score ($D^{low}$) bounding boxes. Any tracklet that is not matched is preserved for a predefined number of frames to test for rebirth in case of occlusion. Finally, the tracklet is removed from the queue if a match is never found.

\section{Experiment and Inference}
\label{experiment}
The development of the Seed Counter using YOLOv8 and object tracking algorithms involves three steps: \newline
\begin{itemize}
    \item Seed detection
    \item Seed tracking
    \item Seed counting
\end{itemize}
\subsection{Seed Detection}
\label{objectdetection}
The YOLOv8 model is trained on the image dataset generated using the synthetic image generator (described in section ~\ref{DR}). 80\% of the image dataset is used for training and 20\% is used for validation. The model is tested on images containing real soy and wheat seed kernels. The test dataset consists of 100 images, 50 of soy and 50 of wheat, each containing 20-30 real seed kernels. Transfer Learning, the ability of a neural network to apply the knowledge gained by training on one dataset to a different dataset where there is a presence of common domains between the source and target datasets, is leveraged to train the YOLOv8 model. Model weights from the YOLOv8 model pretrained on the COCO image dataset are leveraged as provided by Uralytics. The hyperparameters used to train the YOLOv8 model are as shown in Table ~\ref{Hyperparamters}. The results of seed kernel detection on the test dataset are evaluated using the metrics of Precision, Recall, and Average Precision (PR). The metrics are briefly described below and the obtained  results are presented in Table ~\ref{EvaluationResults}. 

\begin{table}[t]
    \centering
    \caption{Hyperparameters for YOLOv8}
    \begin{tabular}{|c|c|}
    \hline
       \textbf{Hyperparameter}  &  \textbf{Value} \\
    \hline
       Learning Rate & 0.001 \\
    \hline
       Batch Size & 32 \\
    \hline
        Input Image Size & 320x320x3 px \\
    \hline
        Bounding Box Confidence Threshold & 0.4 \\
    \hline
        Non-Maximum Suppresion Threshold & 0.4 \\
    \hline
        Intersection-over-Union Threshold & 
        0.5 \\
    \hline 
        Activation Function & LeakyReLU \\
    \hline
        Filters in Each Layer & 64 \\
    \hline
        Dropout & yes \\
    \hline
        Pretrained Model Weights & yolov8x \\
    \hline
    \end{tabular}
    \label{Hyperparamters}
\end{table}

\noindent \textbf{Precision:} Precision indicates the number of positives accurately classified by the classifier from all the classifications made. It is given by \textit{true positives/ (true positives + false positives)}. \newline
\textbf{Recall:} Recall indicates the number of positives identified by the classifier of all the positives present in the dataset. It is given by \textit{true positives/(true positives + false negatives)}.
\textbf{Average Precision:} The Average Precision is computed as the ratio of the Intersection-over-Union (IoU) between the bounding box predictions made by the classifier and ground truth bounding boxes. It is computed for an 50\% overlap between ground truth and predicted bounding boxes for the purposes of this paper, given by $AP_{50}$. \newline 
\textbf{Note:} Average Precision is only reported for the validation data set but not the test data set because images in the test set do not have ground truth bounding boxes plotted around them.  

 \noindent From the results in table ~\ref{EvaluationResults}, high recall scores of 91\% and 90\% on soy and wheat respectively for the test set indicate that the model albeit being trained on synthetic images detects real seed kernels well. The precision scores of 93\% and 92\% indicate that the model classifies the seed kernels correctly in most instances. The reason for high precision might be due to the clear morphometric distinction between soy and wheat.

 \begin{table}[b]
    \centering
    \caption{Evaluation Results}
    \begin{tabular}{|c|c|c|c|c|c|}
    \hline
         \textbf{Seed Kernel}&\multicolumn{3}{|c|}{\textbf{Validation Set}}&\multicolumn{2}{|c|}{\textbf{Test Set}} \\
    \hline
    & Precision & Recall & $AP_{50}$ & Precision & Recall \\
    \hline
    Soy & 98\% & 92\% & 92\% & 93\% & 91\% \\
    \hline
    Wheat & 97\% & 93\% & 89\% & 92\% & 90\% \\
    \hline
    \end{tabular}
    \label{EvaluationResults}
\end{table}

\subsection{Seed Tracking}
\label{SeedTracking}
The Seed Tracking phase involves the application of StrongSORT and ByteTrack algorithms to videos of Soy and Wheat seed kernels captured using the setup described in section ~\ref{HardwareComponents}. The experiment is conducted on videos containing 250 seed kernels of each seed type captured at three different frame rates, 30, 60, and 120 i.e. normal speed, slow motion, and super slow motion. Videos with higher frame rates capture more level of detail than those with lower frame rates. Frame rate has a significant impact on object tracking algorithms since they function by predicting the position of objects in future frames. The StrongSORT and ByteTrack algorithms are applied using the detection weights obtained in the Seed Detection phase as input. The tracking algorithms apply a unique ID to each seed kernel detected in the video and track them throughout the video. In an ideal world where the seed kernels in the video are not prone to occlusion or clustering, mere tracking of objects in the video and counting the number of IDs tracked by the object tracking algorithm is sufficient to obtain a count of the seed kernels in the video. However, the seed kernels in the video are clustered in parts, occluded, and prone to sudden deviations in trajectory, as shown in Fig. ~\ref{WheatSeeds}. The aforementioned issues of clustering, occlusion, and sudden trajectory deviations lead to the risk of object tracking algorithms assigning different unique IDs to the same seed kernel in different frames of the video, eventually leading to a discrepancy between the actual number of seed kernels in the video and number of unique IDs generated by the object tracking algorithms. The number of unique IDs generated by StrongSORT and ByteTrack algorithms on each of the videos captured for the wheat seeds are shown in Tables ~\ref{UniqueIDs} and ~\ref{UniqueIDsByteTrack} respectively. The results show that either of the object tracking algorithms consistently overcount the number of seed kernels in the video.

\begin{table}[t]
    \caption{Unique ID Count on Video of Wheat Seed Kernels by StrongSORT}
    \label{UniqueIDs}
    \centering
    \begin{tabular}{|c|c|c|}
    \hline
        \textbf{Frame Rate} & \ \textbf{Seed Kernel Count} & \textbf{Unique IDs}\\
    \hline
         120 & 250 & 306\\
    \hline
         60 & 250 & 381\\
    \hline
         30 & 250 & 533\\
    \hline
    \end{tabular}
\end{table}

\begin{table}[t]
    \caption{Unique ID Count on Video of Wheat Seed Kernels by ByteTrack}
    \label{UniqueIDsByteTrack}
    \centering
    \begin{tabular}{|c|c|c|}
    \hline
        \textbf{Frame Rate} & \ \textbf{Seed Kernel Count} & \textbf{Unique IDs}\\
    \hline
         120 & 250 & 322\\
    \hline
         60 & 250 & 406\\
    \hline
         30 & 250 & 592\\
    \hline
    \end{tabular}
\end{table}

\subsection{Seed Counting}
\label{SeedCounting}
Seed Counting is performed using a region of interest (RoI) that is established at a common location across each frame in the video. Any seed kernel (track with unique ID) that crosses the RoI is accounted to be one seed kernel. The total number of seed kernels is given by the total number of tracks that cross the RoI. Tables ~\ref{Soy Results} and ~\ref{Wheat Results} show the results obtained by applying the StrongSORT and ByteTrack algorithms on each of the videos for soy and wheat. It is observed that the performance of the object tracking algorithms improves as the frame rate increases. It is perhaps due to the greater detail that is available in videos of higher frame rates. The key issue encountered by the algorithms processing the video is the sudden change in seed kernel trajectory due to the seed kernels colliding with each other and deviating from their current trajectory. This phenomenon affects the object tracking algorithms' ability to predict the location of the seed kernel accurately in subsequent frames. 

\begin{table}[t]
    \centering
    \caption{Results of Soy and Wheat Kernel Count using StrongSORT}
    \begin{tabular}{|c|c|c|c|c|}
    \hline
       \textbf{Seed Type} & \textbf{Frame}  & \textbf{Actual} & \textbf{YOLOv8 } & \textbf{Accuracy} \\
       & \textbf{Rate} & \textbf{Count} & \textbf{Count} & \\
       \hline
        Soy & 120 & 250 & 238 & 95.2 \\
        \hline
        Soy & 60 & 250 & 214 & 85.6 \\
        \hline
        Soy & 30 & 250 & 188 & 75.2 \\
        \hline
        Wheat & 120 & 250 & 233 & 93.2 \\
        \hline
        Wheat & 60 & 250 & 207 & 82.8 \\
        \hline
        Wheat & 30 & 250 & 166 & 66.4 \\
        \hline
    \end{tabular}
    \label{Soy Results}
\end{table}

\begin{table}[t]
    \centering
    \caption{Results of Soy and Wheat Kernel Count using ByteTrack}
    \begin{tabular}{|c|c|c|c|c|}
    \hline
       \textbf{Seed Type} & \textbf{Frame}  & \textbf{Actual} & \textbf{YOLOv8 } &  \textbf{Accuracy} \\
       & \textbf{Rate} & \textbf{Count} & \textbf{Count} & \\
       \hline
        Soy & 120 & 250 & 242 & 96.8 \\
        \hline
        Soy & 60 & 250 & 211 & 84.4 \\
        \hline
        Soy & 30 & 250 & 194 & 77.6 \\
        \hline
        Wheat & 120 & 250 & 231 & 92.4 \\
        \hline
        Wheat & 60 & 250 & 209 & 83.6 \\
        \hline
        Wheat & 30 & 250 & 171 & 68.4 \\
        \hline
    \end{tabular}
    \label{Wheat Results}
\end{table}

From the results in tables ~\ref{Soy Results} and ~\ref{Wheat Results}, it is observed that the seed count is most accurate on videos captured at a frame rate of 120 and least accurate on videos captured at a frame rate of 30 for either seed type. It can be inferred that the performance of the object tracking algorithms directly relates to the frame rate. However, both algorithms consistently undercount the number of seed kernels agnostic of frame rate. It is due to the clustering of seed kernels in certain frames which results in the seed tracking algorithms considering multiple seed kernels as one.

\section{Pitfalls and Future Work}
The key pitfall of the experiment is that the videos used for the experiment consist of seed kernels that are clustered. As a result, the object tracking algorithms fail to track each seed kernel accurately. In future experiments, an augmented image capture setup wherein the seed kernels are not prone to clustering (or occlusion) will be developed. Furthermore, the current image capture setup is feasible only for relatively small seeds. However, a robust image capture technique needs to be developed for larger seed types, such as mango and peach.\newline
\null \quad In the current experiment, the seed kernels of soy and wheat have distinct morphometry which results in the object detector clearly being able to distinguish between the seed types. However, experiments on seed types with morphometry similar to soy, such as canola will need to be performed to determine if the proposed technique applies when seed types of similar morphometry are present. \newline
\null \quad Future work will include the development of a mobile application (Android/iOS) that provides users with the ability to count seed kernels of different seed types. Furthermore, users will be allowed to train the neural network model with their own seed types using the synthetic image generator that will also be made available through the application.

\section{Conclusion}

The experiment demonstrates the feasibility of synthetic images to train object tracking neural network models, and their application in seed kernel counting aimed at the seed packaging industry. Furthermore, the experiment demonstrates yet another application of deep learning for agriculture. The authors are happy to collaborate with anyone interested to pursue the research further. 

\end{document}